\documentclass[11pt,a4paper]{article}

\usepackage[utf8]{inputenc}
\usepackage[T1]{fontenc}
\usepackage{amsmath,amssymb,amsfonts}
\usepackage{graphicx}
\usepackage{booktabs}
\usepackage{hyperref}
\usepackage{xcolor}
\usepackage{algorithm}
\usepackage{algpseudocode}
\usepackage{subcaption}
\usepackage{multirow}
\usepackage[margin=1in]{geometry}
\usepackage{natbib}
\usepackage{enumitem}

\graphicspath{{figures/}}

\hypersetup{
    colorlinks=true,
    linkcolor=blue!70!black,
    citecolor=green!50!black,
    urlcolor=blue!60!black
}

\newcommand{\prism}{\textsc{Prism}}

\title{
\textbf{\textsc{Prism}: Policy Reuse via Interpretable Strategy Mapping}
}

\author{
Thomas Pravetz\\
\texttt{github.com/tompravetz/prism}
}

\date{March 2026}

\begin{document}
\maketitle

% ============================================================
% Abstract
% ============================================================
\begin{abstract}
We present \prism{} (\textbf{P}olicy \textbf{R}euse via \textbf{I}nterpretable \textbf{S}trategy \textbf{M}apping), a framework that grounds reinforcement learning agents' decisions in discrete, causally validated concepts and uses those concepts as a zero-shot transfer interface between agents trained with different algorithms. \prism{} clusters each agent's encoder features into $K$ concepts via K-means. Causal intervention establishes that these concepts directly drive---not merely correlate with---agent behavior: overriding concept assignments changes the selected action in 69.4\% of interventions ($p = 8.6 \times 10^{-86}$, 2500 interventions). Concept importance and usage frequency are dissociated: the most-used concept (C47, 33.0\% frequency) causes only a 9.4\% win-rate drop when ablated, while ablating C16 (15.4\% frequency) collapses win rate from 100\% to 51.8\%. Because concepts causally encode strategy, aligning them via optimal bipartite matching transfers strategic knowledge zero-shot. On Go~7$\times$7 with three independently trained agents, concept transfer achieves 69.5\%\,$\pm$\,3.2\% and 76.4\%\,$\pm$\,3.4\% win rate against a standard engine across the two successful transfer pairs (10 seeds), compared to 3.5\% for a random agent and 9.2\% without alignment. Transfer succeeds when the source policy is strong; geometric alignment quality predicts nothing ($R^2 \approx 0$). The framework is scoped to domains where strategic state is naturally discrete: the identical pipeline on Atari Breakout yields bottleneck policies at random-agent performance, confirming that the Go results reflect a structural property of the domain.
\end{abstract}

\textbf{Keywords:} Reinforcement Learning, Transfer Learning, Concept Bottleneck Models, Interpretability, Knowledge Reuse

% ============================================================
% 1. Introduction
% ============================================================
\section{Introduction}

Reinforcement learning agents embed their strategies in continuous, high-dimensional weight vectors that resist inspection, reuse, or transfer. A PPO agent that masters Go 7$\times$7 over millions of steps cannot share its strategic knowledge with a DQN agent trained on the same task---the two encode their policies in architecturally incompatible forms. Humans face no analogous barrier: a chess player can describe their opening strategy, and another player can adopt it regardless of personal playing style.

The core obstacle is representation. Neural network features are entangled and architecture-specific; there is no obvious common ground between a PPO actor-critic and a DQN Q-network. We propose discrete concept representations as that common ground.

\prism{} clusters each agent's encoder features into $K$ discrete concepts. Because all encoders in a given domain produce fixed-dimensional feature vectors, concepts from different agents live in the same space and can be compared by centroid similarity. Optimal bipartite matching then finds a one-to-one mapping between the two concept sets, and the source policy's embedding table is remapped into the target concept space---zero-shot, no gradient steps required.

Our contributions are:
\begin{enumerate}[leftmargin=*]
    \item \textbf{Causally grounded discrete concepts}: Intervention experiments establish that K-means concepts derived from RL encoder features causally drive agent decisions (69.4\% action-change rate, $p = 8.6 \times 10^{-86}$). Ablation reveals a dissociation between concept frequency and strategic importance: the most-used concept (C47, 33.0\% frequency) causes only a 9.4\% win-rate drop when ablated, while the strategically critical concept (C16, 15.4\% frequency) collapses win rate from 100\% to 51.8\%.
    \item \textbf{Zero-shot cross-algorithm transfer}: Because concepts causally encode strategic decisions, aligning them across agents transfers knowledge zero-shot. PPO$\to$DQN and BC$\to$DQN achieve 69.5\% and 76.4\% win rate, compared to 9.2\% without alignment and 3.5\% for a random agent.
    \item \textbf{Source quality as the dominant factor}: Transfer success is determined by whether the source policy is strong---alignment quality between concept spaces does not predict transfer quality ($R^2 \approx 0$). A weak DQN source fails to transfer to a strong target.
    \item \textbf{Scope characterization}: The identical pipeline on Atari Breakout fails completely (bottleneck at random-agent performance), establishing that PRISM is scoped to domains with naturally discrete strategic structure, and defining this as an empirical boundary condition rather than a design choice.
\end{enumerate}

% ============================================================
% 2. Related Work
% ============================================================
\section{Related Work}

\paragraph{Concept Bottleneck Models.}
Koh et al.~\citep{koh2020concept} introduced CBMs for supervised learning, where intermediate concept predictions---defined by domain experts---are used as a human-readable reasoning layer. Extensions have explored stochastic~\citep{vandenhirtz2024stochastic} and probabilistic~\citep{kim2023probabilistic} variants. Delfosse et al.~\citep{delfosse2024interpretable} applied supervised CBMs to RL using object-centric game features. All of these require human-defined concept labels at training time. \prism{} requires no labels: concepts emerge from K-means clustering of encoder features and are validated post-hoc through intervention, not defined a priori.

\paragraph{Transfer Learning in RL.}
The transfer learning survey of~\citet{taylor2009transfer} identifies inter-task mapping---finding correspondences between state or action spaces across tasks---as a central challenge. Policy distillation~\citep{rusu2016policy} transfers knowledge by training a student to match a teacher's action distribution; progressive networks~\citep{rusu2016progressive} extend active network columns with lateral connections from prior tasks; successor features~\citep{barreto2017successor} separate reward-relevant features from dynamics to enable fast policy switching. These methods transfer at the level of network weights or value functions. \prism{} transfers at the level of discrete symbolic concepts, which are explicitly aligned before transfer and remain interpretable throughout.

\paragraph{Hierarchical and Compositional RL.}
The options framework~\citep{sutton1999between} decomposes policies into reusable sub-policies. HIRO~\citep{nachum2018data} learns goal-conditioned lower-level policies controlled by a higher-level policy. Universal value function approximators~\citep{schaul2015universal} generalize across goal spaces. These methods structure behavior through continuous representations. \prism{}'s discrete concepts provide a coarser but more directly interpretable decomposition that can be shared across independently trained agents.

\paragraph{Discrete Representations in RL.}
World models such as DreamerV2~\citep{hafner2021mastering} use categorical latent variables to support planning. Vector-quantized methods learn discrete codes primarily for reconstruction or compression. \prism{} uses discrete concepts specifically as a transfer interface: the quantization is not an end in itself but a mechanism for producing an alignable representation.

\paragraph{Meta-Learning.}
MAML~\citep{finn2017model} and related approaches optimize for fast adaptation to new tasks via gradient steps. \prism{} requires no gradient updates at transfer time---alignment is computed from centroids and applied directly---making it complementary to meta-learning when the target task is identical to the source.

\paragraph{Unsupervised Skill Discovery.}
DIAYN~\citep{eysenbach2019diversity} discovers diverse behaviors by maximizing mutual information between skills and trajectories. Skills decompose the \emph{action} space into reusable primitives; \prism{} concepts decompose the \emph{state} space into discrete perceptual categories. The two approaches address different parts of the knowledge reuse problem and could in principle be combined.

\paragraph{Causal Interpretability.}
Geiger et al.~\citep{geiger2021causal} formalize interchange interventions to test whether internal representations causally implement a hypothesized computation. Our intervention protocol follows this spirit: we override concept assignments and measure action change to establish causal rather than correlational concept-behavior relationships.

% ============================================================
% 3. Method
% ============================================================
\section{Method}
\label{sec:method}

\subsection{Three-Stage Pipeline}

\prism{} operates in three stages, each producing an artifact that is reused across agents:

\begin{enumerate}[leftmargin=*]
    \item \textbf{Baseline training}: Train a standard RL agent with encoder $f_\theta$. Freeze $f_\theta$ after convergence.
    \item \textbf{Concept discovery}: Collect features $\{f_\theta(o_i)\}$ from gameplay, fit MiniBatch K-means with $K$ clusters, and store the centroid set $\{\mu_k\}_{k=0}^{K-1}$.
    \item \textbf{Bottleneck training}: Train a policy $\pi_\phi$ that receives only the concept assignment $c = \arg\min_k \|f_\theta(o) - \mu_k\|_2$ as input, via an embedding lookup followed by a shallow MLP.
\end{enumerate}

\subsection{Encoder and Bottleneck Architecture}

For Go~7$\times$7, the encoder is a three-layer CNN:
\[
f_\theta(o) = \mathrm{FC}_{128}(\mathrm{ReLU}(\mathrm{Flatten}(\mathrm{Conv}_{64}(\mathrm{Conv}_{64}(\mathrm{Conv}_{32}(o))))))
\]
where $o \in \mathbb{R}^{7 \times 7 \times 3}$ encodes stone positions (black, white, empty). The bottleneck policy maps concept id to action via:
\[
\pi_\phi(a \mid c) = \mathrm{Softmax}(\mathrm{MLP}_{128}(\mathrm{Embed}_{64}(c)))
\]
All three agents (PPO, DQN, BC) use this same encoder and bottleneck architecture, trained independently from random initialization. Because all encoders produce 128-dimensional outputs, their concept centroids are directly comparable.

\subsection{Concept Alignment}
\label{sec:alignment}

Given centroid sets $\{\mu_k^A\}$ and $\{\mu_k^B\}$ from agents $A$ and $B$, we compute the $K \times K$ cosine similarity matrix $S_{ij} = \cos(\mu_i^A, \mu_j^B)$ and find the optimal one-to-one mapping via the Hungarian algorithm~\citep{kuhn1955hungarian}:
\[
\sigma^* = \arg\min_{\sigma \in \mathrm{Perm}(K)} \sum_k -S_{k,\,\sigma(k)}
\]
This runs in $O(K^3)$ time; with $K=64$ it takes under a millisecond.

\subsection{Policy Transfer}
\label{sec:transfer}

The bottleneck policy's embedding table $E \in \mathbb{R}^{K \times d}$ encodes all strategic knowledge in a form that can be remapped. For each target concept slot $j$, we copy the embeddings of the source concepts that map to it:
\[
e_j^B = \frac{1}{|\sigma^{*-1}(j)|} \sum_{i \,:\, \sigma^*(i)=j} e_i^A
\]
and copy the policy MLP weights directly (action spaces match in our experiments). At evaluation time, target agent $B$'s encoder assigns the observation to a concept, and the remapped policy predicts the action---no gradient updates, no fine-tuning.

% ============================================================
% 4. Experimental Setup
% ============================================================
\section{Experimental Setup}
\label{sec:setup}

\paragraph{Environment.} Go~7$\times$7 implemented as a single-agent Gymnasium environment, with $50$ discrete actions (49 board positions plus pass) and $\mathbb{R}^{7\times7\times3}$ observations. We train three independently initialized agents with different algorithms, each from random weights.

\paragraph{Agents.}
\begin{itemize}[leftmargin=*]
    \item \textbf{PPO}~\citep{schulman2017proximal}: MaskablePPO via SB3-contrib, trained via curriculum against GnuGo over 169 generational steps.
    \item \textbf{DQN}~\citep{mnih2015human}: Custom Q-network with action masking, trained via curriculum against GnuGo over 57 generational steps. DQN received substantially less training than PPO, resulting in a weaker policy.
    \item \textbf{BC}: Behavioral cloning using Dataset Aggregation (DAgger)~\citep{ross2011reduction} with GnuGo as the expert, producing a supervised encoder trained on expert demonstrations.
\end{itemize}

All concept managers use $K=64$ and are fitted on features collected from independent gameplay episodes.

\paragraph{Evaluation opponent.} All transfer experiments evaluate against GnuGo, a standard open-source Go engine. GnuGo's strength levels, designed for 19$\times$19 play, do not produce a reliable difficulty gradient on a 7$\times$7 board (a random agent achieves $3.5\% \pm 2.0\%$ win rate against any level from 1 to 10). All experiments use GnuGo as a fixed benchmark opponent; the level parameter should be understood as a configuration constant rather than a difficulty selector.

\paragraph{Statistical protocol.} Main transfer results use 10 seeds $\times$ 100 games per seed. Significance tests are one-sample Welch's $t$-tests against the null hypothesis of 50\% win rate. The true random agent baseline (3.5\%) provides context for interpreting absolute win rates. K-means is initialized with a fixed seed (42) across all experiments; the stability analysis (Section~\ref{sec:validation}) reports NMI~0.587 across random seeds---structurally consistent concept partitions despite label arbitrariness (ARI~0.214). Any replication should fix this seed or apply a stability-selection procedure.

% ============================================================
% 5. Results
% ============================================================
\section{Results}
\label{sec:results}

\subsection{Agent-to-Agent Transfer}
\label{sec:agent_transfer}

Table~\ref{tab:agent_transfer} shows zero-shot transfer results for all six source-target pairs among the three agents. Two transfers succeed decisively: PPO$\to$DQN (69.5\%) and BC$\to$DQN (76.4\%). The remaining four fail to beat GnuGo reliably.

\begin{table}[t]
\centering
\caption{Zero-shot transfer on Go~7$\times$7, 10 seeds $\times$ 100 games. $p$ is a one-sample $t$-test against the 50\% null; random agent achieves 3.5\%. Daggers ($^\dagger$) mark results not significantly above 50\%.}
\label{tab:agent_transfer}
\begin{tabular}{llccc}
\toprule
\textbf{Source} & \textbf{Target} & \textbf{Align Sim} & \textbf{Win Rate} & \textbf{$p$} \\
\midrule
BC  & DQN    & 0.045 & \textbf{76.4\%} $\pm$ 3.4\% & $< 0.001$ \\
PPO & DQN    & 0.021 & \textbf{69.5\%} $\pm$ 3.2\% & $< 0.001$ \\
DQN & PPO    & 0.021 & 49.8\% $\pm$ 2.6\%$^\dagger$ & 0.82 \\
DAgger & PPO & 0.045 & 41.5\% $\pm$ 6.0\%$^\dagger$ & 0.002$^\ddagger$ \\
DQN & BC     & 0.045 & 38.7\% $\pm$ 4.9\%$^\dagger$ & 0.0001$^\ddagger$ \\
PPO & BC     & ---   & 0.0\% (degenerate) & --- \\
\bottomrule
\end{tabular}
{\small $^\ddagger$ Significantly \emph{below} 50\%, but well above the 3.5\% random floor.}
\end{table}

The two successful transfers share a common structure: a capable source (PPO or BC, both achieving $\geq$98\% against GnuGo natively) transfers to DQN as target, whose encoder produces a non-degenerate, functional concept space. The four failing transfers share either a weak source (DQN, 64\% natively, undertrained at 57 vs.\ 169 generations), or a degenerate target encoder.

\paragraph{Degenerate target encoder.} PPO$\to$BC fails with 0\% win rate because the BC encoder collapses all board positions to a single concept during gameplay. This renders the target concept space unusable: regardless of board state, the agent receives the same concept and outputs the same action (pass), ending every game in three moves. The BC agent's own bottleneck achieves 98\% because its policy learned a strong constant-action strategy for that one concept during training---a degenerate but effective solution within its own concept space that breaks down completely when used as a transfer target.

\paragraph{The DQN-as-target pattern.} Both successful transfers share DQN as the target encoder. A reviewer might ask whether this reflects a peculiarity of DQN's concept space---greater regularity, or better alignability---rather than a general mechanism. We cannot rule this out: with three agents in one domain, we do not have an independent replication with a different target. The strongest counter-evidence is DQN$\to$PPO at 49.8\%: a weak DQN source fails to PPO as a target, but this tests a weak source, not a strong one. Whether a fully trained DQN would transfer successfully to PPO is an open question. We report this pattern as a genuine limitation of the current scope.

\paragraph{Alignment similarity does not predict success.} BC$\to$DQN achieves 76.4\% with alignment similarity 0.045, while DQN$\to$BC achieves 38.7\% with the same similarity of 0.045. The similarity is symmetric; transfer quality is not. Source policy strength is the decisive variable.

\paragraph{BC$\to$DQN outperforms PPO$\to$DQN.} The BC agent was trained on GnuGo expert demonstrations---a stronger teacher than what PPO encountered during RL training. Its policy head learned move distributions closer to expert play, and these transfer well to DQN's concept space. PPO's RL-trained policy is also strong but reflects a different distribution of strategic decisions.

\subsection{Alignment Method Comparison}
\label{sec:alignment_comparison}

To determine whether the choice of alignment algorithm matters, we evaluated four methods on the PPO$\to$DQN pair (5 seeds $\times$ 50 games each).

\begin{table}[t]
\centering
\caption{Alignment method comparison for PPO$\to$DQN, 5 seeds $\times$ 50 games. Win rates $\pm$ std. Pairwise Welch's $t$-test vs.\ Hungarian; $^*p < 0.05$. $^\dagger$Greedy NN is degenerate: all source concepts map to one target concept, producing a fixed-action policy (mean game length 48.9 moves vs.\ 27.5 for Hungarian).}
\label{tab:alignment}
\begin{tabular}{lccc}
\toprule
\textbf{Method} & \textbf{Type} & \textbf{Win Rate} & \textbf{$p$ vs.\ Hungarian} \\
\midrule
Greedy NN$^\dagger$      & 1:1 greedy    & 97.2\% $\pm$ 1.6\% & $<$0.001$^*$ \\
Hungarian (PRISM)    & 1:1 optimal   & \textbf{68.8\%} $\pm$ 4.7\% & --- \\
Procrustes           & 1:1 rotated   & 51.6\% $\pm$ 6.4\% & 0.003$^*$ \\
Random permutation   & 1:1 random    & 48.4\% $\pm$ 33.7\% & 0.29 \\
Identity (no alignment) & 1:1 trivial & 9.2\% $\pm$ 3.0\% & $<$0.001$^*$ \\
\bottomrule
\end{tabular}
\end{table}

Four results stand out (Table~\ref{tab:alignment}). First, no alignment (identity mapping) collapses to 9.2\%---near the random agent floor---confirming that alignment is not optional. Second, Hungarian matching outperforms Procrustes and random permutation. Third, random permutation has a mean near 50\% but a standard deviation of 33.7\%: individual seeds range from 0\% (degenerate) to 96\% (lucky permutation). The mean obscures the core finding that random alignment is unreliable---it is a lottery, not a method. Fourth, greedy nearest-neighbor alignment achieves 97.2\%, but this is degenerate: all 64 source concepts map to the same single target concept, so the policy outputs one fixed action regardless of board state. The high win rate reflects a constant-action strategy that happens to work on 7$\times$7; it is not strategic transfer. We include it in Table~\ref{tab:alignment} for completeness.

Procrustes alignment (orthogonal rotation of source centroids before Hungarian matching) performs significantly worse than plain Hungarian (51.6\% vs.\ 68.8\%, $p=0.003$). On the same-task, same-domain setting here, rotating the source centroid cloud distorts the correspondence rather than improving it.

\subsection{Transfer Baselines}
\label{sec:baselines}

We compare PRISM to two reference points on the PPO$\to$DQN pair.

\begin{table}[t]
\centering
\caption{Transfer method comparison, PPO$\to$DQN on Go~7$\times$7.}
\label{tab:baselines}
\begin{tabular}{lcc}
\toprule
\textbf{Method} & \textbf{Zero-Shot WR} & \textbf{Requires Training?} \\
\midrule
\prism{} (10 seeds avg.)    & \textbf{69.5\%} $\pm$ 3.2\% & No \\
From scratch (gen 0)         & 0\%                           & --- \\
From scratch (gen 10, vs.\ GnuGo)  & $\approx$52\%            & Yes (200K steps) \\
From scratch (gen 49 final)  & 72\% (noisy, single seed)     & Yes (1M steps) \\
Random mapping (3 seeds)     & 57\%, 95\%, 0\%               & No \\
\bottomrule
\end{tabular}
\end{table}

\prism{} achieves 69.5\% with no training. A policy trained from scratch on DQN concepts starts at 0\%, reaches approximately 52\% after 10 generations ($\approx$200K steps of training against a random opponent evaluated against GnuGo), and converges to $\approx$72\% after 50 generations---with substantial noise throughout (the learning curve oscillates between 12\% and 80\%).

\begin{figure}[t]
\centering
\includegraphics[width=0.95\linewidth]{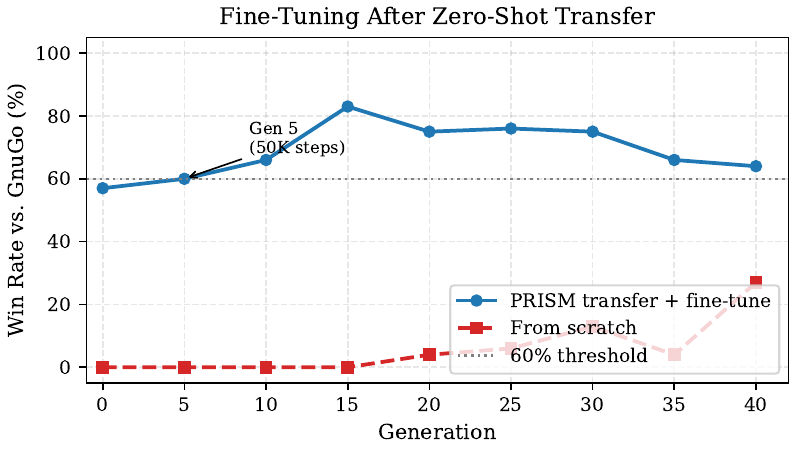}
\caption{Fine-tuning after zero-shot transfer. Both conditions run REINFORCE on the DQN concept bottleneck; they differ only in initialization. The transferred policy crosses 60\% at generation~5 (50K steps); the from-scratch policy reaches 27\% at generation~40 (400K steps) without crossing the threshold. Single seed; treat magnitude as indicative.}
\label{fig:finetune}
\end{figure}

\paragraph{Fine-tuning.}
The zero-shot result is an initialization, not a ceiling. We fine-tuned the transferred policy using REINFORCE on the DQN concept bottleneck and ran an identical REINFORCE training run from a randomly initialized bottleneck (40 generations, 10K steps each, evaluated every 5 generations against GnuGo, 100 games per checkpoint). The transferred policy reaches 60\% win rate at generation~5 (50K steps). The from-scratch policy reaches only 27\% at generation 40 (400K steps) and does not cross the 60\% threshold within this budget. The peak fine-tuned result is 83\% at generation~15 (150K steps). This comparison is internally controlled: both conditions use the same algorithm (REINFORCE on the concept bottleneck) and differ only in initialization. The 8$\times$ step advantage to the 60\% threshold demonstrates that PRISM's zero-shot transfer provides a useful initialization that substantially reduces subsequent RL training cost. This comparison uses a single seed; the direction of the result is clear, but the magnitude should be treated as indicative rather than precise.

\subsection{Concept Validation}
\label{sec:validation}

\paragraph{Interpretability cost.}
Adding the concept bottleneck does not measurably hurt performance (Table~\ref{tab:perf}).

\begin{table}[t]
\centering
\caption{Bottleneck vs.\ unconstrained baseline, mean $\pm$ std win rate aggregated across all GnuGo evaluation runs. GnuGo level has no consistent effect on 7$\times$7; results are pooled.}
\label{tab:perf}
\begin{tabular}{lcc}
\toprule
\textbf{Agent} & \textbf{Bottleneck} & \textbf{Baseline} \\
\midrule
PPO & 99.1\% $\pm$ 0.7\% & 98.5\% $\pm$ 1.5\% \\
DQN & 64.1\% $\pm$ 5.4\% & 54.5\% $\pm$ 12.8\% \\
\bottomrule
\end{tabular}
\end{table}

For PPO the two are within noise. For DQN the bottleneck appears higher, which likely reflects that the unconstrained DQN baseline uses $\epsilon$-greedy exploration at test time (injecting random actions) while the bottleneck uses deterministic argmax. The bottleneck does not outperform---it simply evaluates more consistently.

\paragraph{Causal intervention.}
We test whether concept assignments \emph{cause} action choices by overriding each state's assigned concept with five alternative concepts and measuring how often the chosen action changes (500 states, 5 alternatives each, 2500 interventions total). The action changes in 69.4\% of interventions ($p = 8.6 \times 10^{-86}$, binomial test against the null that concept assignment does not affect action choice). The distribution is heavy-tailed: median per-state change rate is 80\%, but some states are insensitive ($\mathrm{std} = 0.31$), indicating that some board positions constrain the available moves sufficiently that concept identity matters less.

\begin{figure}[t]
\centering
\includegraphics[width=0.95\linewidth]{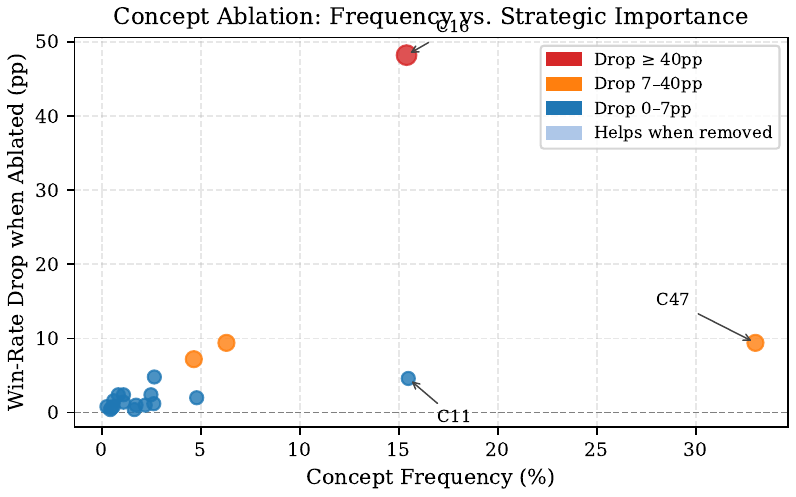}
\caption{Concept ablation: frequency vs.\ win-rate drop when ablated. C16 (15.4\% frequency) causes a 48.2\,pp drop; C47 (33.0\% frequency, the most-used concept) causes only a 9.4\,pp drop. Three concepts improve win rate when removed (negative drop), suggesting they encode suboptimal patterns.}
\label{fig:ablation}
\end{figure}

\paragraph{Concept ablation.}
Table~\ref{tab:ablation} reports the impact of suppressing individual concepts (replacing their action with a uniform random legal move). The baseline bottleneck achieves 100\% win rate against GnuGo; concept C16, active in 15.4\% of states, reduces this to 51.8\% when ablated---near the random agent floor. Three further concepts produce significant but smaller drops (7--9\%). The remaining 60 tested concepts produce drops under 5\%. Two features of this distribution are notable: the most-used concept (C47, 33.0\% frequency) produces only a 9.4\% drop, while the critical concept (C16, 15.4\%) produces a 48.2\% drop---concept frequency and strategic importance are not aligned.

\begin{table}[t]
\centering
\caption{Top concept ablation results. Baseline win rate = 100\% vs.\ GnuGo. Ablation replaces concept actions with uniform random legal moves (500 games per concept).}
\label{tab:ablation}
\begin{tabular}{cccc}
\toprule
\textbf{Concept} & \textbf{Frequency} & \textbf{Ablated WR} & \textbf{WR Drop} \\
\midrule
C16 & 15.4\% & 51.8\% & \textbf{48.2\%} \\
C47 & 33.0\% & 90.6\% & 9.4\% \\
C29 & 6.3\%  & 90.6\% & 9.4\% \\
C6  & 4.6\%  & 92.8\% & 7.2\% \\
\bottomrule
\end{tabular}
\end{table}

\paragraph{Stability.}
Repeating the K-means fit with three different random seeds on the same feature corpus yields a cross-seed ARI of $0.214 \pm 0.009$ and NMI of $0.587 \pm 0.004$. The gap between these two metrics is informative. ARI penalizes any pairwise assignment disagreement, including points near cluster boundaries that land differently across runs. NMI measures the shared information between two partitions, normalized by their entropies, without penalizing relabeling or boundary noise. An NMI of 0.587 indicates that points which cluster together in one initialization tend to cluster together in others: the partition \emph{structure} is substantially shared across seeds even though individual cluster labels are not. This is consistent with K-means finding different orientations of a stable underlying geometric structure in feature space rather than discovering fundamentally different groupings. Perturbation robustness within a single fixed run is 0.999 under Gaussian noise $\sigma{=}0.1$. The practical consequence is that the specific integer label assigned to a concept is an initialization artifact---its strategic content is not---and reported results should be interpreted relative to a fixed trained model.

\subsection{K Sensitivity}
\label{sec:k_sensitivity}

\begin{figure}[t]
\centering
\includegraphics[width=0.95\linewidth]{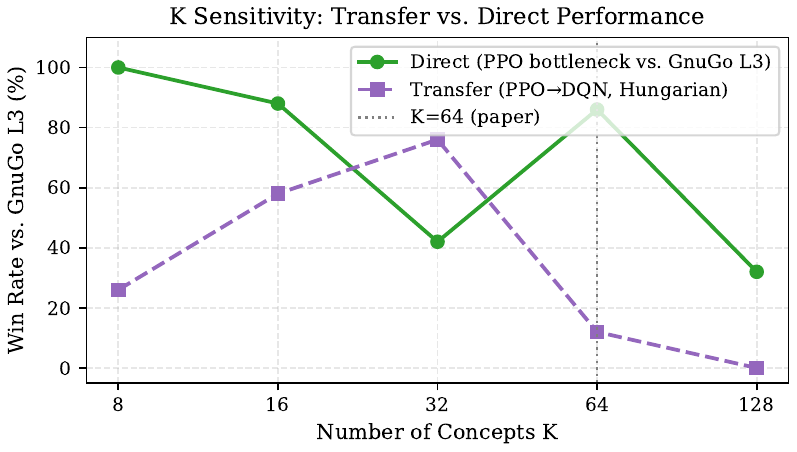}
\caption{Win rate vs.\ $K$ (number of concepts) at 300K training steps against GnuGo L3. Transfer performance peaks at $K{=}32$ (76\%); direct bottleneck performance is higher at $K{=}8$ and $K{=}64$. The paper uses $K{=}64$ based on direct performance at full training (3.4M steps), where larger $K$ captures finer-grained structure. Results at 300K steps are not directly comparable to Table~\ref{tab:agent_transfer} (full curriculum).}
\label{fig:k_sweep}
\end{figure}

To understand sensitivity to the number of concepts $K$, we trained PPO bottlenecks with $K \in \{8, 16, 32, 64, 128\}$ using 300K training steps against GnuGo L3 and measured both direct win rate and PPO$\to$DQN transfer win rate (50 games each). Results are shown in Figure~\ref{fig:k_sweep}.

Transfer win rate follows an inverted-U over $K$, peaking at $K{=}32$ (76\%) and collapsing at $K{=}128$ (0\%). Fewer concepts create a coarser, more generalizable partitioning of state space that aligns well across agents; too many concepts produce a fine-grained decomposition where individual cluster centroids are less reproducible across different training runs, making alignment harder. Direct win rate is highest at $K{=}8$ and $K{=}64$ under the 300K budget, though neither is the best transfer point. The choice of $K$ involves a tradeoff: lower $K$ favors transfer, higher $K$ supports finer-grained direct performance at full training.

The paper uses $K{=}64$ throughout based on its direct performance at the full 3.4M-step curriculum, where the training budget is sufficient to populate all 64 clusters meaningfully. A practitioner targeting transfer rather than direct performance should consider $K{=}32$ or lower.

% ============================================================
% 6. Discussion
% ============================================================
\section{Discussion}
\label{sec:discussion}

\paragraph{When transfer works and when it does not.}
Across the six transfer pairs, two succeed and four fail. The pattern is consistent: transfer requires both a strong source policy and a functional target encoder. These are independently necessary conditions. The strongest source (BC, trained on GnuGo expert demonstrations) fails completely as a target because its encoder maps all observations to one concept. The weakest source (DQN, undertrained) fails to improve any target despite functional encoders on both sides. Alignment quality between concept spaces plays no detectable role once these two conditions are met.

A \emph{concept coverage} account explains why source strength is the operative variable rather than alignment geometry. A source policy trained to convergence encounters the full distribution of strategically distinct board positions, acquiring a meaningful centroid for each of the $K$ concept slots. When aligned to a target, the transferred policy can respond appropriately across the full strategic range. A source that is undertrained has degenerate centroids for positions it rarely or never encountered during training; no alignment can recover information that was never encoded. This coverage account also explains why the alignment similarity metric ($R^2 \approx 0$) predicts nothing: once a source policy has sufficient coverage, its centroid cloud can be aligned to any functional target regardless of their geometric proximity, because the strategic content---not the absolute coordinate position---is what transfers.

\paragraph{The DQN training gap.}
DQN received 57 generational training steps versus 169 for PPO---roughly one-third the compute. Its native win rate ($\approx$64\%) substantially trails PPO ($\approx$99\%). This gap was intentional: we wanted agents of different strength to test whether transfer quality tracks source capability. It does. Had DQN been trained to PPO-level performance, DQN$\to$PPO transfer might succeed; that remains an open question.

\paragraph{Why alignment method matters but alignment quality does not.}
Two findings in Section~\ref{sec:alignment_comparison} appear to be in tension: alignment method matters significantly (Hungarian 68.8\% vs.\ Procrustes 51.6\%, $p=0.003$; vs.\ identity 9.2\%, $p<0.001$), yet the alignment quality metric---normalized centroid similarity after alignment---explains nothing ($R^2 \approx 0$). The resolution lies in what each quantity measures. Alignment \emph{quality} captures geometric proximity: whether the two centroid clouds are close in feature space after alignment. Alignment \emph{method} determines whether strategic pairings are semantically consistent: whether the ``blocking'' concept in the source maps to the ``blocking'' concept in the target. Hungarian matching finds the optimal permutation regardless of global geometry---two agents can have learned the same strategic categories from different initializations, placing their centroid clouds far apart globally, yet still admit a consistent pairwise mapping. Procrustes applies a global rotation first; this helps when the correspondence is a rotation, but hurts when it is a permutation, distorting the pairing. Geometric closeness in feature space is neither necessary nor sufficient for a good strategic correspondence. This is why the method matters (wrong method destroys valid pairings) while the quality metric, which measures something orthogonal to pairing quality, predicts nothing.

\paragraph{GnuGo as a benchmark on 7$\times$7.}
GnuGo's level parameter modulates search depth for 19$\times$19 play. On a 7$\times$7 board the parameter has no consistent effect---a random agent achieves identical win rates against levels 1, 5, and 10. All our results therefore compare against a single fixed opponent, not a difficulty ladder. This limits our ability to characterize agent strength relative to \emph{GnuGo's} capability range, but does not affect any transfer comparison since all experiments use the same opponent configuration.

\paragraph{Scope: Atari Breakout.}
To probe where PRISM's assumptions break down, we applied the full pipeline to a continuous-dynamics domain: Atari Breakout (ALE/Breakout-v5). PPO and DQN agents were trained to competent performance (PPO: 15.1 reward/life; DQN: 9.8 reward/life). K-means concept managers ($K{=}64$) were fit on 512-dimensional NatureCNN features, and bottleneck policies were trained via behavioral cloning using the same procedure as the Go experiments.

The bottleneck policies failed: BC loss flatlined at approximately entropy level (${\approx}1.11$ nats ${\approx}\log 4$ for PPO; ${\approx}0.98$ for DQN) regardless of training epochs, and performance collapsed more than 30$\times$ relative to the baseline (PPO: $15.1\to 0.3$ reward/life; DQN: $9.8\to 0.5$), indistinguishable from the random agent baseline of 0.3 reward/life. Zero-shot concept transfer matched the same floor: PPO$\to$DQN Hungarian $0.4 \pm 0.2$; DQN$\to$PPO Hungarian $0.6 \pm 0.2$ (alignment similarity 0.053). All four conditions---random agent, own bottleneck, Hungarian transfer, identity transfer---cluster within 0.3 reward/life of one another. The concept alignment step adds no signal because the bottleneck itself encodes none.

The failure mode is diagnostic. Breakout requires continuous ball tracking: the correct action (LEFT, RIGHT, or hold) depends on the ball's current velocity and trajectory, which are not recoverable from a cluster assignment over static frame features. K-means groups frames that are geometrically similar in feature space but require opposite actions depending on ball dynamics absent from the centroid representation. The bottleneck therefore learns the marginal action distribution---explaining why BC loss converges to $H(\text{actions})$---rather than a state-conditional policy. This confirms the Go results are not an artifact of the bottleneck architecture but reflect a structural property of the Go domain: distinct board configurations genuinely call for categorically different strategies, making discrete concept assignment sufficient for policy transfer. PRISM is scoped to domains with this property.

\paragraph{Limitations.}
This study uses a single game domain (Go 7$\times$7) with three agents, one of which is degenerate as a transfer target. The transfer results---while statistically robust at 10 seeds---come from a constrained experimental setting. Generalization to larger boards or different game families remains to be shown. Continuous-control tasks where fine-grained dynamics drive action selection appear out of scope for the current K-means bottleneck design, as the Breakout experiment confirms. The K-means stability issue (ARI 0.214 across initializations) means the framework is sensitive to the clustering seed; production use would require either fixing the seed or using a stability-selection procedure. Greedy nearest-neighbor alignment produced degenerate results here; it may behave differently when source and target concept counts differ. The three agents (PPO, DQN, BC) all share the same underlying encoder architecture (GoCNNEncoder) by design; the ``cross-algorithm'' claim refers to training algorithm, not network architecture. Claims of generality to architecturally diverse encoders are untested. DQN's bottleneck win rate (64.1\%) exceeds its unconstrained baseline (54.5\%); we attribute this to $\varepsilon$-greedy exploration degrading baseline performance at evaluation time, but this has not been verified by disabling exploration in the baseline. Finally, the BC bottleneck's 98\% win rate with a single-concept (constant-action) representation suggests GnuGo on 7$\times$7 can be exploited by consistent play; this is consistent with our disclosure that GnuGo's level parameter has no consistent effect on 7$\times$7, and does not affect any transfer comparison, but it does mean the absolute win-rate scale should be interpreted accordingly.

% ============================================================
% 7. Conclusion
% ============================================================
\section{Conclusion}
\label{sec:conclusion}

\prism{} establishes discrete concept representations as a practical transfer interface between reinforcement learning agents. On Go 7$\times$7, Hungarian alignment of K-means concept spaces enables zero-shot policy transfer achieving 69.5\% and 76.4\% win rate against a standard engine---with no gradient updates at transfer time, and no measurable performance cost from the concept bottleneck itself. The intervention and ablation experiments confirm that concepts causally drive behavior and that the concept space is not uniform: a single concept (C16) accounts for nearly half the agent's winning capability when absent.

The central finding is that transfer quality depends on source policy strength, not on the geometric alignment between concept spaces. This has a practical implication: when selecting a source agent for transfer, optimize for policy quality, not for concept similarity to the target.

The concept bottleneck structure extends naturally beyond machine-to-machine transfer. Because concepts causally drive behavior and are represented as fixed-dimensional embeddings, they can in principle be paired with natural language descriptions --- moving from a system where concepts are shared between agents to one where they are explained to humans. This would shift the desideratum from geometric alignability to linguistic describability, a different and complementary research direction.

% ============================================================
% References
% ============================================================
\bibliographystyle{plainnat}
\bibliography{prism}

\end{document}